\documentclass[sigconf,natbib=true]{acmart}

\usepackage{caption}
\usepackage[T1]{fontenc}

\usepackage[utf8]{inputenc}

\usepackage{microtype}

\newcommand{\questgen}{\textsc{QuestGen}}
\begin{document}
\title{\questgen: Effectiveness of Question Generation Methods for Fact-Checking Applications}

 \author{Ritvik Setty$^*$}
  \email{ritvik.setty@gmail.com}
 \affiliation{
   \institution{Fremont High School}
   \city{Sunnyvale}
   \state{CA}
   \country{USA}
 }

 \author{Vinay Setty}
  \email{vsetty@acm.org}
 \affiliation{
   \institution{University of Stavanger}
   \city{Stavanger}
   \country{Norway}
 }
 
\additionalaffiliation{%
   \institution{Factiverse AI}
   \city{Stavanger}
   \country{Norway}}

\begin{abstract}
Verifying fact-checking claims poses a significant challenge, even for humans. Recent approaches have demonstrated that decomposing claims into relevant questions to gather evidence enhances the efficiency of the fact-checking process. In this paper, we provide empirical evidence showing that this question decomposition can be effectively automated. We demonstrate that smaller generative models, fine-tuned for the question generation task using data augmentation from various datasets, outperform large language models by up to 8\%. Surprisingly, in some cases, the evidence retrieved using machine-generated questions proves to be significantly more effective for fact-checking than that obtained from human-written questions. We also perform manual evaluation of the decomposed questions to assess the quality of the questions generated.
\end{abstract}


\maketitle
\section{Introduction}
\label{sec:intro}
As a response to growing online misinformation, and factual mistakes resulting from large language models (LLMs) the significance of online fact-checking has surged, prompting the development of automated methodologies~\cite{chen2023combating}. Consequently, scalable, effective fact-checking mechanisms has become critical. Automated fact-checking systems aim to streamline and enhance this process, leveraging advanced technologies to manage the overwhelming amount of information online. The standard framework for automated fact-checking comprises a tripartite pipeline~\cite{Gao:2023:arXiv}: (1) Identification of Check-worthy claims \cite{hassan2017toward}, (2) Web-based searches to gather Supporting or Refuting evidence  and (3) Assessment of the claim's veracity based on the retrieved evidence~\cite{popat2018declare,augenstein2019multifc,Setty:2024:SIGIR}. In this paper, our aim is to apply question generation techniques to improve the performance of the last two stages.

The pipeline works well for straightforward claims that can be verified with a single document, such as ``The population of the US is 334 million'', which can be checked using a Wikipedia page. However, it struggles with complex claims containing multiple sub-claims that require diverse documents for thorough evaluation~\cite{ousidhoum2022varifocal,yang2022explainable,Chern:2023:arXiv}. For instance, verifying ``President Joe Biden stated that unemployment has been below 4\% for the longest stretch in over 50 years'' requires breaking it into specific queries about historical unemployment trends and similar periods, necessitating the aggregation and reasoning of various data sources.

In this regard, recent works have released human-annotated datasets for claim decomposition into questions. Studies by \cite{Fan:2020:arXiv,Chen:2022:arXiv,Schlichtkrull:2023:NEURIPS,Pan:2023:ACL,Krishna:2022:TACL} have contributed valuable resources and methodologies for this purpose. These datasets provide a foundation for developing systems that can effectively break down complex claims into manageable components, facilitating more accurate and comprehensive fact-checking. In this paper, we compile a comprehensive list of datasets available for decomposing fact-checking claims into questions. In addition, we also introduce two synthetic datasets: (1) \textit{Repurposing a QA Dataset}: We adapt an existing question-answering (QA) dataset \cite{Kwiatkowski:2019:TACL} for the claim decomposition task similar to \citet{Park:2022:arXiv} who adapt it for the fact verification task (FavIQ). This involves reformatting and recontextualizing the data to suit the needs of fact-checking, ensuring that the questions generated are relevant and useful for verification purposes. (2) \textit{Generating Synthetic Questions Using the GPT-3.5-Turbo Model}: Leveraging the capabilities of the GPT-3.5-Turbo model~\cite{ouyang2022training}, we generate synthetic questions to supplement existing datasets. This approach allows us to create a diverse and extensive collection of questions, enhancing the robustness and effectiveness of automated fact-checking systems.

Our goal is to compare the performance of fine-tuned smaller models and larger models under one-shot chain of though prompt settings. By evaluating the efficacy of different models, we aim to identify the most effective approaches for automated claim decomposition and verification. We also show how we can leverage transfer learning to help improve performance on other datasets. Transfer learning involves using pre-trained models on one dataset to improve performance on another, facilitating knowledge transfer and enhancing overall system capabilities.

We formulate the following research questions:

\begin{itemize}

    \item RQ1: Can we automate question generation from the claims for fact-checking? How does it compare to human-written questions? 

    \item RQ2: Can data augmentation improve the performance of question generation?

    \item RQ3: How well do the generated questions help improve the performance of retrieval and claim verification? 

    \item RQ4: Can smaller models like BART and T5 learn to generate based on data generated from larger models like GPT-3.5? 

\end{itemize}

Our contributions in summary are answers to these research questions and introducing a benchmark for the question generation task for fact-checking. We believe these contributions are valuable to the field of automated fact-checking and development of more effective tools and methodologies. We also release the code and datasets used for this paper \url{https://github.com/factiverse/questgen}.

\begin{figure*}[ht!]
    \centering
    \small
    \begin{minipage}{0.55\textwidth}
        \centering
        \includegraphics[width=\textwidth]{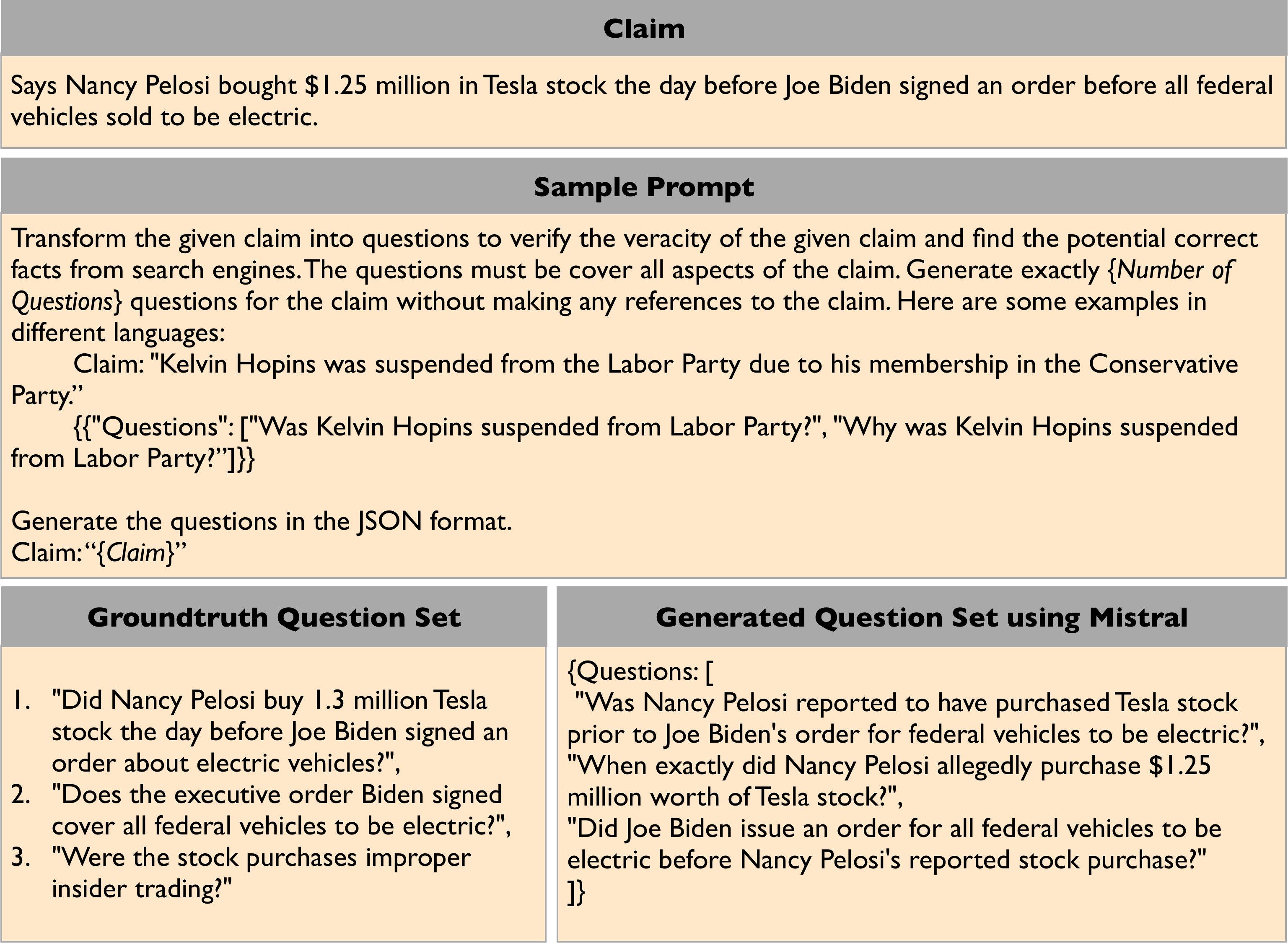}
        \caption{Prompt, claim, and questions generated using Mistral.}
        \label{fig:example}
    \end{minipage}\hfill
    \begin{minipage}{0.42\textwidth}
        \centering
        \captionof{table}{Dataset statistics.}
        \label{tab:data-statistics}
        \scalebox{0.95}{
        \begin{tabular}{p{2cm}|p{1cm}p{1cm}p{1cm}p{1cm}}
        \hline
        \textbf{Dataset} & \textbf{\# claims} & \textbf{Avg. \# questions} & \textbf{Training size} & \textbf{Testing size} \\ 
        \hline
        ClaimDecomp+\cite{Chen:2022:arXiv}  & 100 & 2.85 & 0 & 285 \\
        ClaimDecomp\cite{Chen:2022:arXiv}  & 5,465 & 1.00 & 4,400 & 1,088 \\
        QABriefs \cite{Fan:2020:arXiv} & 7,353 & 3.16 & 18,281 & 1,456 \\
        AVeriTeC \cite{Schlichtkrull:2023:NEURIPS} & 3,486 & 2.66 & 7,985 & 1,287 \\
        favIQ~\cite{Park:2022:arXiv} & 146,162 & 1.00 & 140,977 & 5,877 \\
        GPT-3.5-gen & 62,954 & 3.03 & 152,716 & 38,179 \\
        \hline
        \end{tabular}
        }
        
        \vspace{0.5cm} %
        \captionof{table}{Manual evaluation of question generation methods (Usefulness, Coverage and Fluency). Inter-annotator agreement is measured with weighted $\kappa$.}
        \label{tab:ManEval}
        \scalebox{0.95}{
        \begin{tabular}{l|ccc|c}
        \hline
        \textbf{Model} & \textbf{Usefulness} & \textbf{Coverage} & \textbf{Fluency} & \textbf{Weighted $\kappa$} \\ \hline
        BART & 2.8083 & 3.4667 & 4.1750 & 0.2312 \\
        Flan-T5 & 1.5167 & 4.1250 & 3.1583 & 0.2179\\
        T5-base & 3.2750 & 3.7500 & 4.5417 & 0.4287\\
        Mistral-7b & \textbf{4.1667} & \textbf{4.2157} & \textbf{4.6961} & \textbf{0.6514}\\
        \hline
        \end{tabular}
        }
    \end{minipage}
\end{figure*}

\section{Methodology}
\label{sec:method}
In this section, we describe our process for question generation.

\subsection{Question generation methodology}
\label{sec:method:One-Shot}
\paragraph{One-shot Chain of Thought}
For LLMs, we design a one-shot prompt with chain of thought (CoT) reasoning. See Figure \ref{fig:example} for an example. The LLMs generate questions in JSON format; non-conforming responses are counted as null and scored accordingly.

\paragraph{Nucleus Sampling}
\label{sec:method:Nucleus}

To generate multiple unique questions for a claim, we use nucleus sampling \cite{Holtzman:2019:arxiv}, which samples from the most likely tokens in a model's predictions. Nucleus sampling depends on: (1) temperature: emphasizes higher probability tokens,  (2)  top$_k$: restricts choices to the $k$ tokens with maximum scores, and (3) top$_p$: ensures the selected tokens' cumulative probability is $\leq p$.

\subsection{Claim Verification}
\label{sec:method:ClaimVerif}
After generating questions for a claim, the next step is to search for highly relevant supporting and refuting articles to verify the claims. The quality of the generated questions is evaluated based on their effectiveness for fact-checking.

We use search engines, including Google, Bing, You.com, Wikipedia, and Semantic Scholar. Fact-checker websites are avoided to prevent ground truth leakage. Both the claim and generated questions are used as queries, and the top 20 relevant snippets are selected using cosine similarity on sentence embeddings. These claim, snippet pairs are then passed to a natural language inference model (\textit{XLM-RoBERTa-Large}) trained on synthetic data (FEVER) and past fact-checks from professional fact-checkers (see \cite{Setty:2024:SIGIR} for more details). The predictions of claim, snippet pairs are aggregated using majority voting (similar to \cite{popat2018declare,Botnevik:2020:SIGIR,Setty:2024:SIGIR}) and a final label is assigned.

\section{Experimental Evaluation}
In this section, we describe the setup for training SLMs and taking inferences from SLMs and LLMs. We will then analyze the results.

\subsection{Datasets}
\label{sec:method:Datasets}
As shown in Table \ref{tab:data-statistics}, we use five datasets for the question generation task. All datasets contain multiple questions per claim. For training, each question is paired with a claim to create a distinct input-output (claim-question) pair. While some pairings may contain the same input (claim), the input-output (claim-question) pair is unique. We use three human-annotated datasets (\textsf{AVeriTec}, \textsf{ClaimDecomp}, \textsf{QABriefs}) and two synthetic datasets (\textsf{favIQ}, GPT-3.5-gen). The \textsf{favIQ} dataset,  derived from a QA dataset \cite{Kwiatkowski:2019:TACL}, is repurposed for question generation similar to favIQ\footnote{favIQ does not leverage the questions themselves for the fact verificiation task.}. We also collected 62,954 claims from fact-checkers via the Google Fact Check API\footnote{\url{https://developers.google.com/fact-check/tools/api}} and generated questions using a prompt in Figure \ref{fig:example}.

\subsection{Models}
\label{sec:method:Models}
For the task of question generation, we utilize two categories of models: Small Language Models (SLMs) and Large Language Models (LLMs). The SLMs, including Flan-T5-Base, T5-Base, and BART-Base, are fine-tuned on specific datasets. SLMs are fine-tuned to generate questions given a claim. For ablation, the models are fine-tuned on each dataset, and all the datasets combined.  We consider open LLMs Mistral-7b and Llama2-7b for large-scale evaluation. But we also consider GPT-4 for the smaller scale evaluation for the claim verification. When training the models with combined datasets, we introduce them in the order of their size and complexity akin to curriculum learning.

We fine-tune all SLMs as Seq2Seq models from the HuggingFace framework\footnote{\url{https://huggingface.co/}}. The max input and output tokens is 1024 tokens. The models are trained for 10 epochs on a batch size of 8 with the AdamW optimizer and an initial learning rate of $2e-5$. Weight decay is set to $0.3$. SLMs are fine-tuned on a machine with $4$ T4 GPUs or $1$ A100 GPU with CUDA. Questions are generated from two classes of models: Small Language Models (SLMs) and Large Language Models (LLMs). For nucleus sampling in SLMs, we set top$_p$ of $0.95$, top$_k$ of 40 and temperature of $1.5$. Other parameters for SLM inference are tfs$_{z}$ of 1.0, max prediction tokens of $300$ and repetition penalty of $1.1$. For LLMs, we use the Ollama framework to deploy Mistral and Llama2-7b and use the default parameters.

\subsection{Metrics}
\label{sec:method:Metrics}
We use BLEU, Rouge 1 and Rouge L scores to evaluate and compare accuracy values between models. Given a claim-question pair $(c,q)$ from a test set $d$ and a model $M$, we generate a question $M(c) \to g$ as described in Section \ref{sec:method:One-Shot} and Section \ref{sec:method:Nucleus}. For a metric $m$, the metric score $m(q,g)$ is computed. The final score for metric $m$ on model $M$ with dataset $d$ is computed as the arithmetic mean of $m(q,g)$ for all $(c,q)$, $M(c) \to g$ in the dataset $d$. We also use \textbf{Manual Evaluation} to manually evaluate generated questions using human judgements evaluating usefulness, coverage, and fluency.

\begin{table*}[ht!!!]
\small
    \centering
        \caption{Question generation evaluation for different datasets, models fine-tuned by combining train splits from respective datasets and using LLMs. $\dagger$ indicates the results are statistically significant with $p\leq0.05$ using the student's t-test.}
    \scalebox{0.9}{
    \begin{tabular}{lcccccccccccccccc}
    \toprule
     \textbf{Model}&\multicolumn{3}{c}{\textbf{AvertiTec}} & \multicolumn{3}{c}{\textbf{ClaimDecomp}} & \multicolumn{3}{c}{\textbf{QABriefs}} & \multicolumn{3}{c}{\textbf{FavIQ-R}}& \multicolumn{3}{c}{\textbf{GPT-3.5-gen}}  \\
   &\multicolumn{1}{c|}{R-1} & \multicolumn{1}{c|}{R-L} & \multicolumn{1}{c}{BLEU}&\multicolumn{1}{c|}{R-1} & \multicolumn{1}{c|}{R-L} & \multicolumn{1}{c}{BLEU}&\multicolumn{1}{c|}{R-1} & \multicolumn{1}{c|}{R-L} & \multicolumn{1}{c}{BLEU}&\multicolumn{1}{c|}{R-1} & \multicolumn{1}{c|}{R-L} & \multicolumn{1}{c}{BLEU} & \multicolumn{1}{c|}{R-1} & \multicolumn{1}{c|}{R-L} & \multicolumn{1}{c}{BLEU}  \\
     \midrule
     \multicolumn{3}{l}{\textbf{Fine-tuned w/ single dataset}}\\
     Flan-T5  &  0.2593 & 0.2318 & 0.0578 &  0.3841 & 0.3384 & 0.1554 & 0.281 & 0.246 & 0.0628 & 0.5639 & 0.534 & 0.2349 & 0.246 & 0.2203 & 0.0418 \\

T5-base & 0.0006 & 0.0006 & 0.0000  & 0.0252 & 0.0251 & 0.0001 & 0.3044 & 0.2736 & 0.0979 & 0.585 & 0.5791 & 0.5232 &0.1332 & 0.1209 & 0.027 \\
BART & \textbf{0.3324} &\textbf{0.309} & \textbf{0.1237}& 0.8014 & 0.7604 & 0.5503  & \textbf{0.3903} & \textbf{0.3517} & \textbf{0.1372} & 0.9094 & 0.9007 & 0.8057 & \textbf{0.4305} & \textbf{0.3898} & \textbf{0.1328} \\
\midrule
\multicolumn{3}{l}{\textbf{Fine-tuned w/ all datasets}}\\
Flan-T5  &  0.2593 & 0.2318 & 0.0578 &  0.3841 & 0.3384 & 0.1554 & 0.281 & 0.246 & 0.0628 & 0.5639 & 0.534 & 0.2349 & 0.246 & 0.2203 & 0.0418 \\

T5-base & 0.3117 & 0.2798 & 0.0751 & 0.7468 & 0.7147 & 0.5559 & 0.3231 & 0.2994 & 0.0882 & \textbf{0.9257}$^\dagger$ & \textbf{0.9183}$^\dagger$ & \textbf{0.8127}$^\dagger$& 0.4199 & 0.3782 & 0.1225 \\
BART & 0.3111 & 0.2811 & 0.0861 & \textbf{0.8230}$^\dagger$ & \textbf{0.7977}$^\dagger$ & \textbf{0.6395}$^\dagger$  & 0.3745 & 0.3476 & 0.1125 & 0.9123 & 0.9051 & 0.7792 & 0.4256 & 0.3818 & 0.1302 \\
\midrule
  \multicolumn{3}{l}{\textbf{LLM (1-shot CoT)}} \\
Llama2-7b &  0.1490 & 0.1464 & 0.0033 & 0.8018 & 0.7684 & 0.5134 & 0.2731 & 0.2335 & 0.0218 & 0.5311 & 0.4967 & 0.0989 &  0.2287 & 0.1989 & 0.0234 \\
Mistral-7b &  0.2800 & 0.2414 & 0.0380 & 0.813 & 0.7736 & 0.5438 & 0.2793 & 0.2337 & 0.0236 & 0.6685 & 0.6336 & 0.1617  & 0.3128 & 0.2658 & 0.0372 \\
\hline
\label{tab:performance}
     \end{tabular}
     }
     \end{table*}
\section{Results}

\subsection{Automated Evaluation}
\label{sec:results}
\begin{table}[ht!!]
\centering
\caption{Evaluation of veracity prediction using RoBERTa-Large NLI model using  question generation methods.}
\scalebox{0.9}{
\begin{tabular}{lcccc}
\hline
\textbf{Method} & \textbf{Macro F1} & \textbf{Micro F1} & \textbf{True F1} & \textbf{False F1} \\ \hline
\textbf{Claim only}    & 0.4879 & 0.6098 & 0.7377 & 0.2381 \\
\midrule
\textbf{Human written}    &  0.5536 & 0.5600 & 0.6071 & 0.5000 \\
\midrule
\multicolumn{3}{l}{\textbf{Fine-tuned models}}\\
BART      & 0.5895 & 0.6000 & 0.6552 & 0.5238  \\
Flan-T5   & 0.5464 & 0.5900 & 0.6870 & 0.4058 \\
T5-base        & \textbf{0.6250} & \textbf{0.6400} & \textbf{0.7000} & \textbf{0.5500} \\
\midrule
\multicolumn{3}{l}{\textbf{LLM (1-shot CoT)}}\\
Mistral-7b & 0.6161 & 0.6300 & 0.6891 & 0.5432 \\
LLama2-7b & 0.5769 & 0.6061 & 0.6880 & 0.4658 \\
GPT-4 & 0.5746 & 0.5900 & 0.6554 & 0.4938 \\ \hline

\end{tabular}
}
\label{tab:nli}
\end{table}

In this section, we answer \textit{RQ1: Can we automate question generation for fact-checking?} based on the automated evaluation metrics.

As shown in Table \ref{tab:performance}, BART consistently outperforms other models across various datasets. In the \textsc{AvertiTec} dataset, BART achieves the highest scores when fine-tuned with a single dataset and performs well with all datasets. For the \textsc{ClaimDecomp} and \textsc{QABriefs} datasets, BART excels in both ROUGE and BLEU metrics. In the \textsc{FavIQ-R} dataset, T5 achieves the highest scores when fine-tuned with all datasets, nearing human-level performance with ROUGE-1 of 0.9257, ROUGE-L of 0.9183, and BLEU of 0.8127, while BART also performs exceptionally well. For the \textsc{GPT-3.5-gen} dataset, BART remains the top performer, demonstrating its robustness with the highest ROUGE and BLEU scores. Overall, fine-tuning models on specific datasets significantly improves the quality of generated questions, with BART and T5 approaching human-level performance in the FavIQ-R dataset.

One interesting observation is that when training language models like T5-base exclusively on individual datasets such as \textsc{AvertiTec}, its performance is suboptimal. However, when the training data is augmented by combining multiple datasets, the model shows significant improvements in performance. We also notice that BART results are statistically significant when trained on all datasets but not when trained on single datasets.

The key insight here is that the large-scale datasets, specifically GPT-3.5-gen and FavIQ-R, which each have around 150k training examples, contribute most significantly to this improvement. This finding suggests that the \textbf{volume and diversity of training data play a crucial role in enhancing the model's ability to generate questions effectively}. Therefore, the answer to \textit{RQ2} is indeed Yes: data augmentation does improve the performance of question generation, provided that the datasets involved are large-scale.

\subsection{Manual Evaluation}
n this section, we address RQ1 using manual evaluation, as automated metrics may not accurately reflect model performance. Due to the high cost of manual evaluation, we selected BART, Flan-T5, and T5-base, known for their strong performance, and included Mistral-7b for comparison with Llama-7b. Two evaluators assessed each generated question for usefulness, coverage, and fluency on a scale from 1-5, using a subset of 20 claims from the ClaimDecomp+ set with 58 questions.The final score, the arithmetic mean of the evaluators' scores, is listed in Table \ref{tab:ManEval}. The evaluation focused on the quality of questions based on their usefulness, coverage, and fluency. Mistral-7b received the highest ratings, with a weighted $\kappa$ of 0.6514, indicating substantial agreement among evaluators. T5-Base was also competitive, with a weighted $\kappa$ of 0.4287, indicating moderate agreement. For annotation instructions, see: \footnote{ \url{https://github.com/factiverse/questgen/blob/develop/doc/eval.md}}

\subsection{Downstream NLI Task evaluation}
Next we answer RQ3 by evaluating the evidence retrieval and claim verification tasks jointly on a smaller dataset, \textsc{ClaimDecomp+} proposed in \cite{Chen:2022:arXiv}. In this experiment, we evaluate four different categories of question generation methods: (1) Claim only: Using the claim text to retrieve evidence, (2) Human written: Questions written by humans, (3) Fine-tuned models: Generative models fine-tuned on all training datasets, and (4) LLM (1-shot CoT): LLMs using a prompt with one example and reasoning.

For evidence retrieval, the `claim only' method uses only the claim text, while the other methods use both the claim text and the generated questions. To ensure a fair comparison, we limit the evidence to the top 20 most relevant snippets for all methods. We then use a Roberta-Large NLI model to predict the stance of the claim-evidence pairs as True (Supporting) or False (Refuting).

As shown in Table \ref{tab:nli}, generated questions improve the overall performance of the claim verification task. Among the question generation methods, SLMs, specifically T5-base, outperform all other methods, with LLMs like GPT-4 and Mistral-7b being close. This answers our RQ4 that \textbf{smaller models like Mistral-7b and T5-base can outperform GPT-4 when fine-tuned}. Interestingly, human-written questions perform poorly compared to those generated by models, possibly because humans tend to focus on more nuanced questions rather than covering all aspects of the claim.

\subsection{Discussion}
\label{sec:analysis}
Automating the generation of questions from claims for the purpose of fact-checking is a practical approach that can be effectively handled by smaller language models. However, it is important to highlight that the quality of the generated questions, as measured by automated metrics or through manual evaluation, may not necessarily correlate with their performance in the downstream NLI task. Despite this, manual evaluations have shown consistency with NLI performance results, indicating that Mistral-7b and T5-Base models are the best-performing solutions. These models have even demonstrated superior performance compared to GPT-4.
\section{Related work}
\label{sec:related}
Fact-checking is a complex process even for humans\footnote{\url{https://www.niemanlab.org/2024/01/asking-people-to-do-the-research-on-fake-news-stories-makes-them-seem-more-believable-not-less/}} and professional fact-checkers spend hours to days in verifying claims \cite{Gao:2023:arXiv}. Automation or assisting humans in fact-checking has received a lot of attention in the literature recently~\cite{Setty:2024:SIGIR,Setty:2024:SIGIRa,augenstein2019multifc,popat2018declare,Hu:2024:AAAI,V:2024:SIGIR}. However, complex claims require deep analysis of the information needed for verification. Using claim alone does not give sufficient context for machines to retrieve the necessary evidence to verify the claims. Therefore, several methods have been proposed in the literature for generating questions for verifying claims~\cite{Fan:2020:arXiv,Chen:2022:arXiv,Schlichtkrull:2023:NEURIPS,Krishna:2022:TACL,Pan:2023:ACL}. For example, \citet{Fan:2020:arXiv} generate the dataset (\textsc{QABriefs}) using existing claims from other datasets and generating questions using (\textsc{QABriefer}) based on BART trained on \textsc{QABreifs} Dataset. Following this, \citet{Chen:2022:arXiv} propose ClaimDecomp, decomposition of claims and justification into implied and literal questions in the form of yes-no questions, which the authors claim are more comprehensive and precise. The authors also include 100 example literal and implied questions as a representative example of ClaimDecomp (which we call ClaimDecomp+) for evaluations. \citet{Park:2022:arXiv} proposes a large fact-checking dataset based on a QA dataset, FAVIQ from ambiguous QA questions. Following this, \citet{Schlichtkrull:2023:NEURIPS} propose AVERITEC, a QA dataset which uses real-world claims with evidence retrieved from the web. We omit methods which use program verification~\cite{Pan:2023:ACL} and theorem proving \cite{Krishna:2022:TACL}, since the questions they generate are fundamentally different. In this paper, we focus on generating questions for complex claims from multiple datasets using multiple models. We show the efficacy of smaller language models to produce results similar to large language models. To the best of our knowledge, there are no existing works which comprehensively evaluate fact-checking systems under a large-scale benchmark using question generation.

\section{Conclusion}
\label{sec:concl}
In this paper, we demonstrate that question generation for fact-checking applications can be automated using sequence-to-sequence generative models, such as T5, BART, and large language models like Mistral and Llama. We introduce a large-scale benchmark that includes both synthetic and human-written questions corresponding to claims. By employing both manual and automated evaluation metrics, we confirm that it is indeed feasible to effectively automate question generation techniques. Additionally, we assess the downstream NLI task to illustrate the effectiveness of question generation for fact-checking purposes.

\section{Acknowledgements}
This work is in part funded by the Research Council of Norway project EXPLAIN (grant number 337133).
\newpage
\balance
\bibliography{acl2024-questgen}
\bibliographystyle{ACM-Reference-Format}
\end{document}